\documentclass[letterpaper]{article}
\usepackage{aaai}
\usepackage{times}
\usepackage{helvet}
\usepackage{courier}
\usepackage{natbib}
\usepackage{graphicx}
\usepackage{float} 
\usepackage{amssymb}
\usepackage{amsmath}
\usepackage{booktabs}
\usepackage[hidelinks]{hyperref}

\nocopyright
 
\begin{document}

\title{Breaking the Barriers: Video Vision Transformers for Word-Level Sign Language Recognition}

\author{Alexander Brettmann, Jakob Grävinghoff, Marlene Rüschoff, Marie Westhues\\
University of Cologne\\
Albert-Magnus-Platz \\
50923 Cologne\\
jgraevin@smail.uni-koeln.de
}
\date{\today}
\maketitle
\begin{abstract}
\begin{quote}
Sign language is a fundamental means of communication for the deaf and hard-of-hearing (DHH) community, enabling nuanced expression through gestures, facial expressions, and body movements. Despite its critical role in facilitating interaction within the DHH population, significant barriers persist due to the limited fluency in sign language among the hearing population. Overcoming this communication gap through automatic sign language recognition (SLR) remains a challenge, particularly at a dynamic word-level, where temporal and spatial dependencies must be effectively recognized. While Convolutional Neural Networks (CNNs) have shown potential in SLR, they are computationally intensive and have difficulties in capturing global temporal dependencies between video sequences. To address these limitations, we propose a Video Vision Transformer (ViViT) model for word-level American Sign Language (ASL) recognition. Transformer models make use of self-attention mechanisms to effectively capture global relationships across spatial and temporal dimensions, which makes them suitable for complex gesture recognition tasks. 
The VideoMAE model achieves a Top-1 accuracy of 75.58\% on the WLASL100 dataset, highlighting its strong performance compared to traditional CNNs with 65.89\%. Our study demonstrates that transformer-based architectures have great potential to advance SLR, overcome communication barriers and promote the inclusion of DHH individuals.

\end{quote}
\end{abstract}

\section{Introduction}
\noindent
Human interaction relies on language. Through a combinations of words, gestures, and vocal tones our emotions, desires, and personality can be expressed in several settings. For those experiencing profound hearing loss, sign language emerges as the indispensable primary means of communication \citep{alaghband2023survey}. Worldwide, more than 1.5 billion people are affected by speech or hearing loss. This number is expected to rise to 2.5 billion by 2050, of whom 700 million will require care \citep{who_hearing_loss}. Among the deaf and hard-of-hearing (DHH) population, over 70 million individuals depend on sign language for daily communication \citep{earthweb_sign_language_users}. However, substantial barriers persist due to the limited fluency in sign language among the hearing population, thereby impeding daily interactions \citep{kothadiya2022}.

This communication gap has far-reaching implications across social, healthcare, and economic domains. Socially, DHH individuals often experience exclusion and isolation. In healthcare settings, barriers in communication lead to dissatisfaction and avoidance of medical services, adversely affecting health outcomes \citep{rannefeld2023deaf, rogers2024health}. Economically, unaddressed hearing loss incurs an annual global cost of approximately \$980 billion, stemming from productivity losses, healthcare expenses, and disparities in education and employment opportunities \citep{who_hearing_loss}. Addressing these challenges through effective recognition and translation of sign language into written words offers a scalable and impactful solution to bridge this communication divide.

Sign language is a sophisticated form of communication that combines manual signals (e.g. hand shape, movement) to convey words and sentences as well as non-manual signals (e.g. facial expressions, body movements) to communicate grammatical and emotional meaning \citep{alaghband2023survey, elakkiya2021optimized}. It shows distinct syntax, structure, and grammar compared to spoken language with over 135 unique regional variations \citep{nidcd_asl}. Furthermore, sign language can be categorized into static and dynamic forms. Static signs involve fixed hand and facial gestures, while dynamic signs include both isolated gestures, representing a single word, and continuous sequences of gesture that form complete sentences \citep{kothadiya2022}. This study focuses on dynamic and isolated signs, which are particularly challenging due to their reliance on temporal sequences and subtle gesture variations. These inherent complexities of sign language, such as contextual variations in sign meaning, the need to identify subtle differences, distinguish similar signs, and interpret intricate temporal sequences, present substantial challenges for automated recognition systems \citep{elakkiya2021optimized, kothadiya2022, li2020word}.

Historically, SLR has leveraged image-based and video-based approaches, with our focus on video-based methods. Convolutional Neural Networks (CNNs) and their variants, such as 2D-CNN, 3D-CNN, and hybrid CNN-RNN models, have been applied to extract spatial and temporal features from video data, demonstrating notable success \citep{kumari2024, li2020word, kishore2018selfie, shin2019korean, pigou2015sign, ye2018recognizing, huang2018videobasedsignlanguagerecognition}. While these deep learning models have advanced the field, they are often limited in capturing long-range dependencies and require substantial computational resources. Moreover, existing video-based SLR models struggle with challenges such as cluttered backgrounds, varying illumination, and limited dataset sizes, which hinder their generalization and scalability \citep{li2020word}.

In this context, our project introduces a novel approach using ViViT for word-level SLR. Unlike CNN-based models, ViViTs utilize self-attention mechanisms to capture global spatial and temporal relationships within video sequences,  enhancing the model's ability to recognize subtle differences in sign gestures \citep{Arnab_2021}. Utilizing a subset of the large-scale WLASL2000 dataset, which comprises over 21,000 videos across more than 2,000 ASL words, we fine-tune the pre-trained ViViT models, TimeSformer \citep{gberta_2021_ICML} and VideoMAE Transformer \citep{tong2022videomae}, on the WLASL100 subset to ensure computational feasibility. We also incorporate data augmentation techniques to improve model generalization. By benchmarking a ViViT-based approach against state-of-the-art CNN models like I3D \citep{li2020word}, this research aims to demonstrate the potential of transformer-based architectures to advance video-based SLR, ultimately contributing to breaking down communication barriers and promoting inclusion for DHH individuals.

\section{Related Work}

Several studies have examined SLR using deep learning methods, with models trained and evaluated on various datasets. Many focus on image-based models using datasets such as the ASL dataset, Sign Language MNIST, and Indian Sign Language (ISL), which contain static images of ASL or ISL alphabets \citep{goswami2021, barbhuiya2021,bantupalli2018, sharma2021}. In contrast, fewer studies target video-based models, which leverage public datasets like the American Sign Language Lexicon Video Dataset (ASLLVD), Word-Level American Sign Language (WLASL), and IISL2020 for dynamic ASL recognition \citep{kumari2024,bantupalli2018,kothadiya2022}. Video-level models are more relevant for capturing the full complexity and context of ASL gestures, which often depend on motion and temporal sequences, and are therefore the focus of our work. 

Existing models primarily rely on CNNs to process spatial and temporal data from videos. One example is the Inflated 3D ConvNet (I3D) from \cite{li2020word}, which uses 3D convolutions to capture both spatial and temporal features. Other popular CNN architectures such as Alexnet, VGG16, and MobileNetV2 have also been modified and applied for this purpose \citep{bantupalli2018, kumari2024, sharma2021}. In addition, to process the sequential information present in video data, Recurrent Neural Networks have been integrated into CNN models. \cite{li2020word} employed such a hybrid approach in which spatial features extracted from a VGG16 were further processed by a Gated Recurrent Unit (GRU) to model the temporal dynamics of human pose keypoints. Similarly, \cite{kothadiya2022} combined Long Short-Term Memory (LSTM) and GRU layers to effectively handle temporal dependencies, allowing the model to capture and identify gesture sequences from video input. In addition, attention mechanisms have been incorporated to account for significant signs in video sequences. For example, \cite{kumari2024} have proposed an attention-based hybrid CNN-LSTM model that extends the model's ability to focus on key temporal features, which further improves the accuracy and robustness of SLR systems.

Although CNN-based models have been shown to be effective, they are limited in their ability to capture broad temporal dependencies and tend to be computationally intensive. In contrast, ViViT models have been shown to be effective for a variety of visual tasks \citep{Khan2022, han2022survey, akbari2021vatt}. They are very suitable for video data due to their self-attention mechanisms that capture wide-ranging dependencies and allow the model to focus on significant signs in a sequence \citep{Arnab_2021}. This makes them ideal for tasks such as SLR. Despite its enormous potential, research on ViViT models for video-based SLR is still limited. Therefore, our project aims to fill this gap by developing a ViViT model for dynamic ASL recognition.

\section{Methodology}
To investigate the impact of transformer-based architectures on sign language video data recognition, we employ two pre-trained transformer models: TimeSformer \citep{gberta_2021_ICML} and VideoMAE \citep{tong2022videomae} and compare them to the fine-tuned I3D model of \cite{li2020word}.

\begin{figure}[ht]
    \centering
    \includegraphics[width=1\linewidth]{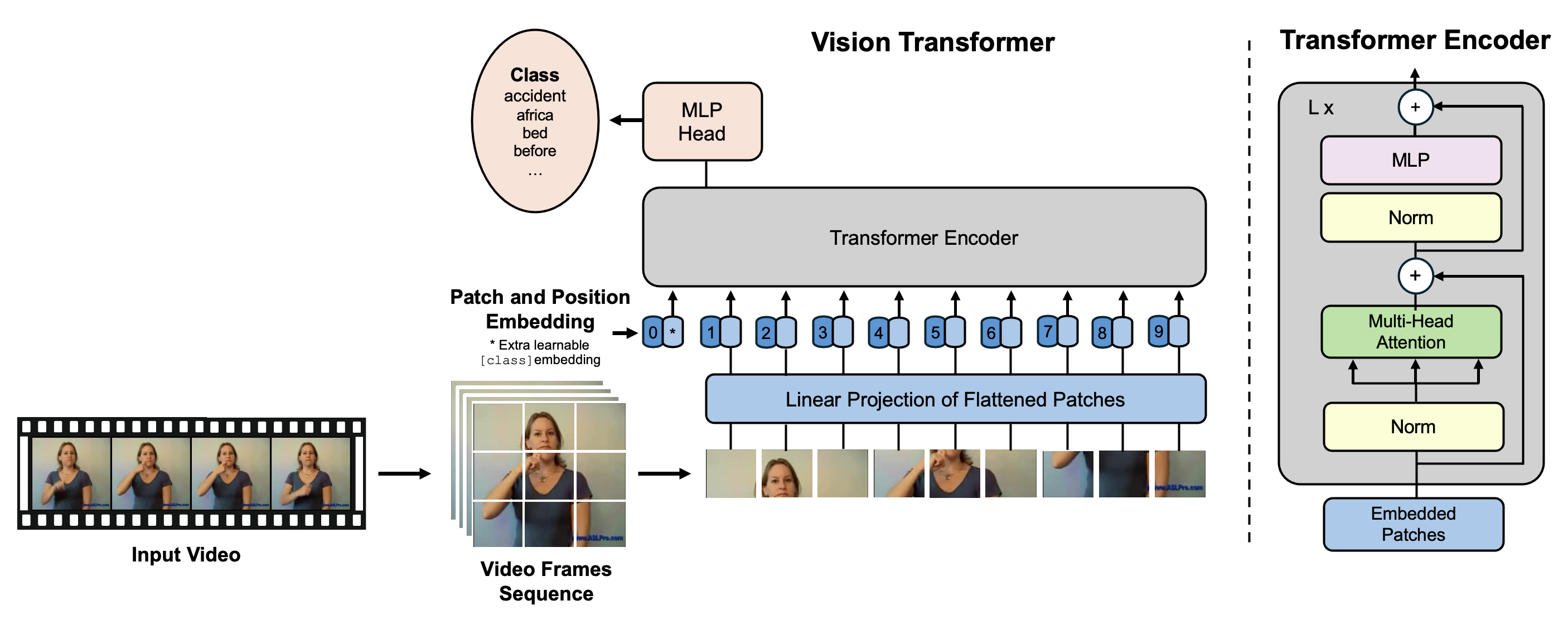}
    \caption{ViViT Architecture. Adapted from \cite{DBLP:journals/corr/abs-2010-11929}}
    \label{fig:pipeline}
\end{figure}

The TimeSformer model of \cite{gberta_2021_ICML} introduces a convolution-free approach to video classification by leveraging self-attention mechanisms. It builds upon the Vision Transformer architecture \citep{DBLP:journals/corr/abs-2010-11929}, illustrated in Figure \ref{fig:pipeline}, by extending it to handle video sequences. This is achieved by encoding both spatial and temporal relationships within a transformer-based framework. The input to the TimeSformer is a video clip represented as a 3D tensor 
$X \in \mathbb{R}^{H \times W \times 3 \times F}$, where F is the number of frames, and each frame has spatial dimensions $H \times W$ with three color channels (RGB). Each frame is divided into non-overlapping patches, flattened, and projected into a feature space using a learnable linear embedding. These feature vectors are augmented with positional embeddings that encode their spatiotemporal locations, ensuring the model retains the order and relationships between patches across both dimensions. Self-attention is applied along the temporal dimension to capture dependencies across frames. This enables the model to understand the evolution of visual elements over time and effectively capture motion and temporal context. Additionally, self-attention is applied within each frame across spatial patches, allowing the model to learn spatial dependencies such as object structure, shape, and intricate inter-pixel relationships. The TimeSformer separates spatial and temporal attention into distinct operations within each transformer block. This divided space-time attention strategy enhances computational efficiency while maintaining the model's ability to capture both dependencies.
The TimeSformer model is constructed using multiple stacked transformer blocks, each designed to capture spatiotemporal dependencies in video data effectively. Each transformer block comprises two main components: a multi-head self-attention mechanism and a feed-forward neural network. The multi-head self-attention mechanism operates either on the spatial or temporal dimensions, enabling the model to focus on critical spatial features within individual frames or temporal patterns across frames. Complementing this, the feed-forward neural network enhances the expressiveness of the model while maintaining stability during training through the use of residual connections and layer normalization. Additionally, the TimeSformer employs a classification token, which is a special learnable token prepended to the sequence of patch embeddings. This token interacts with other embeddings throughout the transformer layers, gathering global spatiotemporal information. The final representation of the classification token, after passing through all transformer blocks, serves as the video-level feature and is utilized for classification tasks.

The specific model employed in our study was initially pre-trained on the extensive ImageNet-21K dataset \citep{deng2009imagenet} to learn robust visual representations and was subsequently fine-tuned on the Kinetics-400 dataset \citep{Carreira_2017_CVPR} for action recognition tasks. For our WLASL100 dataset, we fine-tuned the last three layers of the model over 15 epochs, utilizing a batch size of 4.

\begin{figure}[ht]
    \centering
    \includegraphics[width=1\linewidth]{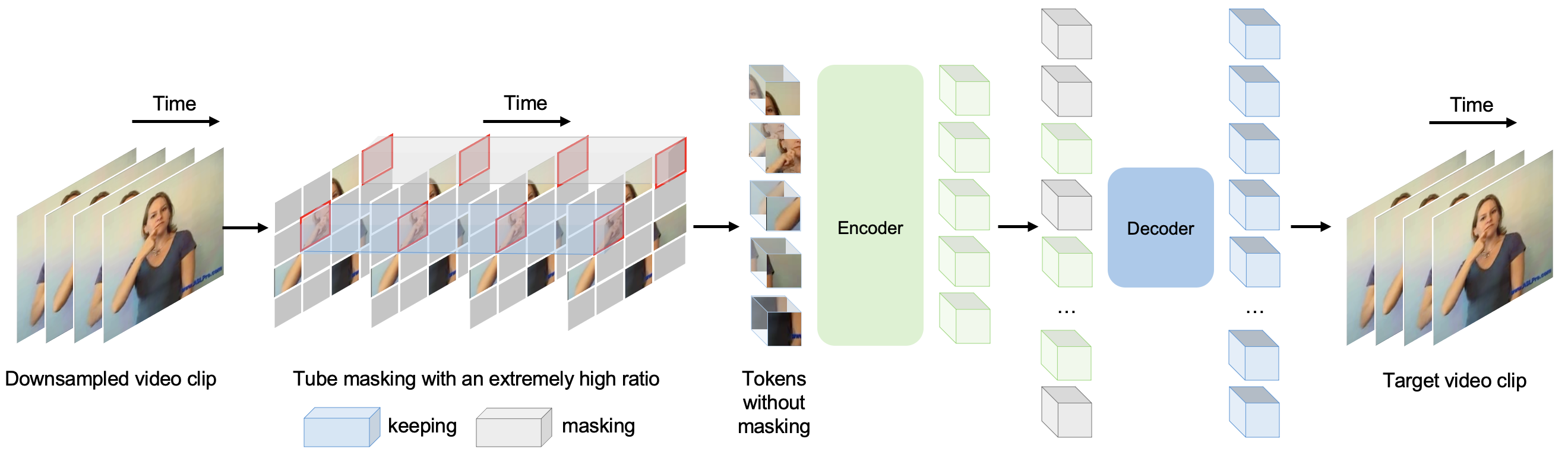}
    \caption{Illustration of VideoMAE. Adapted from \cite{tong2022videomae}.}
    \label{fig:VideoMAE}
\end{figure}

The VideoMAE model, as proposed by \cite{tong2022videomae}, is based on the principles of masked autoencoders (MAE), which are designed to handle video data by reconstructing hidden segments. This method encourages the model to learn meaningful representations by challenging it to fill in masked parts of the input \citep{He_2022_CVPR}. 
A key aspect of VideoMAE is its use of a high masking ratio, between 90\% and 95\%. This technique utilizes the natural repetition found in video data, improving the model's pre-training performance and reducing computational demands due to its efficient encoder-decoder structure shown in Figure \ref{fig:VideoMAE}.
The model employs a tube masking strategy that applies a uniform masking pattern across multiple video frames. This helps the model to understand larger semantic relationships, beyond just immediate frame-to-frame connections, and prevents information leakage in sections of minimal motion. This approach is supported by the Vanilla Vision Transformer backbone \citep{DBLP:journals/corr/abs-2010-11929}, enhanced with joint space-time attention to better capture complex spatiotemporal dynamics \citep{Arnab_2021}. VideoMAE also incorporates temporal downsampling to increase efficiency. By reducing the number of frames processed while retaining crucial visual information, the model can focus on key temporal features without overwhelming data.
Additionally, VideoMAE employs a joint space-time cube embedding strategy, with each cube of size \(2 \times 16 \times 16\) functioning as a token embedding \citep{Arnab_2021}. Here, \(T\) is the temporal dimension (frames), and \(H\) and \(W\) represent each frame's height and width. The embedding layer produces \(\frac{T}{2} \times \frac{H}{16} \times \frac{W}{16}\) 3D tokens, mapping them to a channel dimension \(D\). This approach enables detailed analysis of spatial and temporal dynamics, essential for understanding video content. The architecture includes an asymmetric encoder-decoder design. The encoder focuses on visible video parts, while the lightweight decoder reconstructs the hidden segments, optimizing resource usage, and improving the model's ability to interpret visible data effectively.

As highlighted in the work of \cite{tong2022videomae}, the VideoMAE model demonstrates effective performance on small datasets, making it particularly well-suited for our WLASL100 dataset. Our dataset comprises only about 2,000 videos, yet VideoMAE effectively harnesses this limited labeled data to demonstrate its capabilities. We have shown that its efficiency makes it suitable for situations with restricted data availability, emphasizing its strong performance in accurately recognizing and interpreting complex sign gestures.

The proposed transformer-based approaches are benchmarked against the I3D model applied by \cite{li2020word}, which performs very well on the video-based SLR task using the WLASL dataset. The I3D model, first introduced by \cite{Carreira_2017_CVPR}, is a deep learning architecture designed for video classification tasks. The model adapts traditional 2D CNNs by inflating their convolutional and pooling filters from 2D (spatial) to 3D (spatiotemporal). This allows both spatial and temporal features to be extracted from video data simultaneously. Its backbone is based on the Inception-v1 architecture (2D CNN), a common pre-trained image classification architecture with batch normalization. By inflating the 2D filters of Inception-v1 into 3D, the I3D model captures the spatial representation of individual frames such as height and width and the temporal relationships across frames in a hierarchical manner. This design enables the model to effectively capture dynamic motions, such as hand movements and arm orientations, which are crucial for distinguishing between signs. 

The original inflated 3D model of \cite{Carreira_2017_CVPR} was pre-trained on ImageNet \citep{russakovsky2015imagenet}, a large-scale image dataset, and then fine-tuned on the Kinetics-400 dataset \citep{Carreira_2017_CVPR}, which consists of diverse action recognition videos. To represent sign language specific features, such as hand shapes or facial expressions, \cite{li2020word} initially trained the entire I3D model on the WLASL2000 dataset. They then fine-tuned the pre-trained model on smaller subsets of the WLASL dataset (e.g. WLASL100, WLASL300) by replacing the original classification layer of the I3D model with a new fully connected layer corresponding to the number of classes in the subset. \cite{li2020word} followed the original training configurations of \cite{Carreira_2017_CVPR} which involved using 64 consecutive frames from each video as input. Each input frame was resized to a spatial resolution of 224×224 pixels and RGB videos were processed using 3 input channels. In addition, all models were trained using the Adam optimizer and 200 epochs on each subset. The training was stopped when the validation accuracy stopped to increase. 

While the I3D model remains a strong basis for video-based recognition tasks, it is computationally complex and limited in capturing long-range dependencies due to the local nature of convolutional operations. Transformer-based models like TimeSformer and Video MAE address these limitations and challenges of dynamic SLR by applying global attention mechanisms and efficient pre-training strategies. 

\section{Experiment}
\begin{figure}[ht]
    \centering
    \includegraphics[width=1\linewidth]{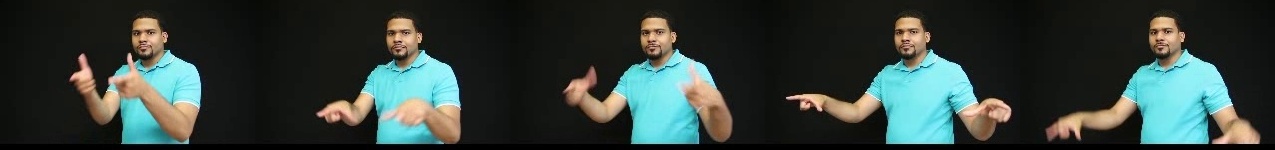}
    \caption{Frames for Sign "Language"}
    \label{fig:sign-lanuage}
\end{figure}

In our project, we use the WLASL dataset \citep{li2020word} which is a large-scale video dataset for word-level SLR. It consists of over 2,000 different words with gestures performed by multiple signers in various environments.
For computational reasons, we focus on a subset, WLASL100, which contains the 100 most frequent words. In this context, a gloss, meaning the written label that corresponds to a specific sign, is assigned to each word. Each gloss is associated with several video instances that demonstrate the corresponding sign in different contexts, with a total of 2,038 videos. The amount of videos per gloss ranges between 18 and 40 with a median of 20 videos per gloss. The videos have an average of 62 frames, with a minimum of 12 and a maximum of 203 frames. As an example, the sign "language" is shown with 5 sample frames in Figure \ref{fig:sign-lanuage}. For the training and evaluation of the model, the samples of a gloss are divided into training, test, and validation sets with a ration of 4:1:1 corresponding to the split applied by \cite{li2020word}. Since we aim to compare our results with those reported by \cite{li2020word}, who combined the training and validation sets for training and used the test set as both validation and test sets, we follow the same approach to ensure consistency in evaluation.

\begin{figure}[ht]
    \centering
    \includegraphics[width=0.6\linewidth]{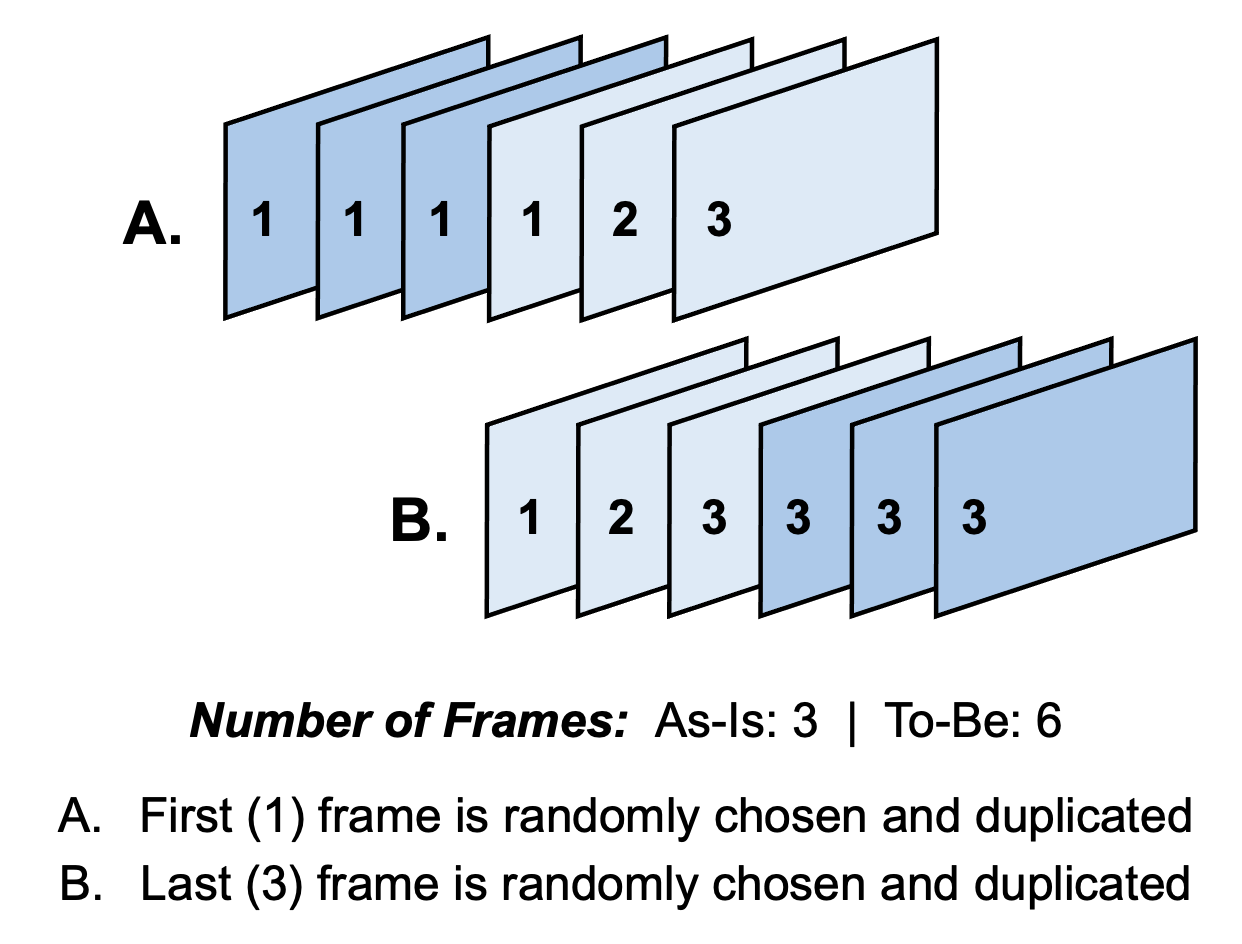}
    \caption{ Padding}
    \label{fig:Padding}
\end{figure}

Before using the videos for training, they are pre-processed in a similar way to \cite{li2020word}. First, to ensure that all videos contain the same number of frames, a random starting point is selected within the video, and the target number of consecutive frames is extracted from that point. To compare this approach, we also explored an alternative pre-processing strategy where frames are sampled evenly throughout the video. If a video contains fewer frames than the target, padding is applied. This means that the first or last frame is randomly chosen and duplicated repeatedly until the desired number of frames is reached (see example in Figure \ref{fig:Padding}). In addition, the resolution of all video images is adjusted to ensure consistent dimensions for model input. If the smaller dimension is less than 226 pixels, it is scaled up to 226 pixels. If the larger dimension is greater than 256 pixels, it is scaled down to 256 pixels. Further, the frames are converted from BGR to RGB to meet the model's expected format. 

After extracting and resizing the frames, they are further augmented. In the set we use for training, a random patch of 224x224 is extracted from each video input frame and then flipped horizontally with a probability of 0.5. The extraction and flipping operations are consistently applied to all frames of a video in the same manner, ensuring uniform pre-processing across the entire video rather than treating each frame independently. In the test set, 224x224 patches are extracted from the center of the frame and no flipping is applied. Once the data was pre-processed, we fine-tuned the TimeSformer and VideoMAE models on the prepared dataset.

To assess the model’s performance, validation/test accuracy is used as main evaluation metric. In the context of our 100-class classification task, accuracy is calculated by dividing the number of correctly classified samples by the total number of samples, since each sample belongs to exactly one class:

\begin{equation}
\text{Accuracy} = \frac{Correct\ Predictions}{Total\ Predictions}
\end{equation}
\\
More specifically, we are using top-\( K \) accuracy with \( K = \{1, 5, 10\} \). Since similar gestures are often used for different meanings, it can lead to errors in classification. However, by incorporating contextual information, some of these misclassifications can be corrected \citep{li2020word}. Therefore, relying on top-\( K \) accuracy is a more effective approach  for SLR at  word level.

\begin{table}[h!]
\centering
\renewcommand{\arraystretch}{1.5}
\begin{tabular}{lccc}
\toprule
\toprule
\textbf{Accuracy}         & \multicolumn{3}{c}{\textbf{Model}} \\ 
\midrule
\midrule
                 & TimeSformer & VideoMAE & I3D$^*$ \\ 
\midrule
top-1            & 62.02\%       & \text{75.58\%}    & 65.89\% \\ 
\hline
top-5            & 87.98\%       & \text{91.86\% }   & 84.11\% \\ 
\hline
top-10           & 94.19\%      & \text{95.74\%}    & 89.92\% \\ 
\bottomrule
\bottomrule
\end{tabular}
\vspace{0.2cm}
\newline \footnotesize $^*$ Results from \cite{li2020word}. 
\caption{Top-\( K \) accuracy (\%) of TimeSformer, VideoMAE, and I3D Model.}
\label{tab:accuracy}
\end{table}

The results shown in Table \ref{tab:accuracy} demonstrate that the VideoMAE model outperforms both TimeSformer and I3D on the WLASL100 dataset, achieving a top-1 accuracy of 75.58\%, compared to 62.02\% for TimeSformer and 65.89\% for I3D. Similarly, VideoMAE excels in top-5 and top-10 accuracy, scoring 91.86\% and 95.74\%, respectively, surpassing the results of TimeSformer (87.98\% and 94.19\%) and I3D (84.11\% and 89.92\%).
Notably, the I3D model used by \cite{li2020word} was pre-trained on the WLASL2000 dataset, which contains more than 20,000 videos. They fine-tuned all layers on WLASL2000 and subsequently retrained only the classification layer for the smaller WLASL100 dataset with roughly 2,000 videos. In contrast, we trained all layers of the VideoMAE model directly on the WLASL100 dataset, utilizing significantly fewer epochs than \cite{li2020word} despite achieving better performance.
This comparison highlights the efficiency of VideoMAE, which demonstrates strong performance without requiring extensive pretraining on larger datasets or significant computational resources for fine-tuning. It underscores the model's capacity to effectively learn both spatial and temporal patterns in sign language videos directly from a smaller dataset like WLASL100.

\begin{table}[ht]
\centering
\resizebox{\linewidth}{!}{%
\begin{tabular}{cccccccc}
\hline
\textbf{Batch} & \textbf{Epochs} & \textbf{Frames} & \textbf{Init. LR} & \textbf{Model} & \textbf{Fine-Tuned Layers} & \textbf{Sampling} & \textbf{Top-1 Acc. (\%)} \\ \hline
4              & 20       & 64              & $1e{-4}$          & TimeSformer    & 3    &  Consec.   & 61.24      \\ 
4              & 20       & 16              & $1e{-5}$          & TimeSformer    & 12   & Even    & 60.85         \\
4              & 15       & 16              & $1e{-5}$          & TimeSformer    & 3    &  Even    & 62.02         \\ 
6              & 20       & 16              & $1e{-5}$ & VideoMAE    & 12  & Consec.     & 46.12    \\
6              & 20       & 16              & $1e{-5}$ & VideoMAE    & 12  & Even       & 62.02       \\
6              & 30       & 16              & $1e{-5}$ & VideoMAE    & 12  & Even       & 75.58       \\
6              & 30       & 16              & $1e{-5}$ & VideoMAE    & 12  & Even       & 66.28$^*$    \\   \hline
\end{tabular}%
}
\vspace{0.2cm}
\caption{Ablation Study Results}
\footnotesize
\begin{minipage}{\linewidth}
\emph{Notes:} 
$^*$ Indicates models trained using only the training set. Validation performance was used for model selection, and the final model was evaluated on the test set.
\textbf{Init. LR}: Initial learning rate used for training. 
\textbf{Sampling}: Consec. refers to randomly sampling consecutive frames, while Even refers to sampling frames evenly distributed across the video.
\end{minipage}
\label{tab:training_results_finetuned}
\end{table}

In addition, we conducted an ablation study to evaluate the impact of hyperparameters, such as batch size and the number of fine-tuned layers, on model performance (Table \ref{tab:training_results_finetuned}). Our results also indicate that the frame sampling strategy significantly impacts model performance. The TimeSformer model using 16 evenly spaced frames outperforms the model trained with 64 consecutive frames when fine-tuning only the last three layers. This suggests that sampling evenly distributed frames better captures the overall temporal dynamics of sign language videos, leading to improved recognition performance.

\begin{figure}
    \centering
    \includegraphics[width=0.9\linewidth]{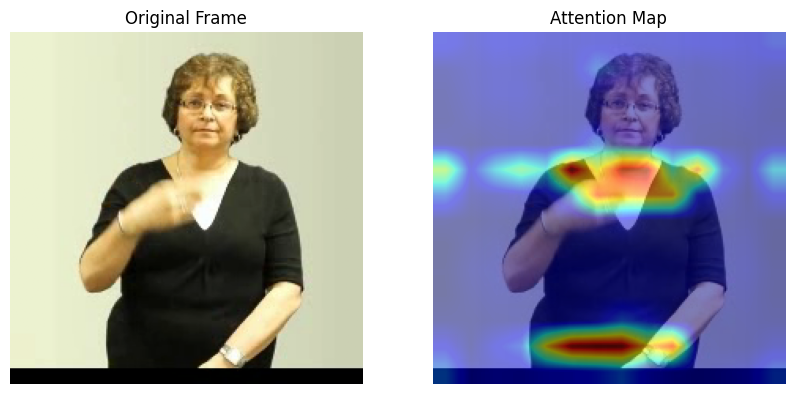}
    \caption{ASL under Attention Map}
    \label{fig:AttentionMap}
\end{figure}

A crucial component of ViViT models is their attention mechanism, which is applied across both spatial and temporal dimensions to extract meaningful features from video frames. This process is illustrated in Figure \ref{fig:AttentionMap}, where the left panel shows the original video frame, and the right panel visualizes the attention map generated by the model. The attention map highlights the regions of the frame that the model focuses on, particularly the hands and upper body of the individual, which are essential for recognizing sign language gestures. This ability to selectively attend to relevant areas make ViViTs highly effective for video-based tasks.

\section{Conclusion}
\noindent Our research demonstrates the potential of transformer-based models, such as VideoMAE and TimeSformer, for advancing video-based SLR. Despite using significantly fewer resources and epochs compared to prior CNN-based models, VideoMAE achieves superior accuracy on the WLASL100 dataset, outperforming both TimeSformer and the fine-tuned I3D model. These results highlight the efficiency of VideoMAE in capturing complex spatiotemporal patterns, marking a step forward in bridging communication barriers for the DHH community. Future work can focus on extending this approach to larger datasets (such as WLASL2000) and moving from word-level to sentence-level recognition. Moreover, further model optimizations and other transformer architectures could be investigated in the future to enhance performance.

\begin{quote}
\begin{small}
\clearpage
\bibliography{reference.bib}
\bibliographystyle{aaai}
\end{small}
\end{quote}
\end{document}